%% file: Main.tex
\documentclass[conference]{IEEEtran}
\IEEEoverridecommandlockouts
\usepackage{cite}
\usepackage{amsmath,amssymb,amsfonts}
\usepackage{algorithmic}
\usepackage{graphicx}
\usepackage{textcomp}
\usepackage{xcolor}
\usepackage[numbers]{natbib}
\usepackage{booktabs}
\usepackage{url}
\usepackage{color,soul}
\def\BibTeX{{\rm B\kern-.05em{\sc i\kern-.025em b}\kern-.08em
    T\kern-.1667em\lower.7ex\hbox{E}\kern-.125emX}}
\begin{document}

\title{DeepTimeAnomalyViz: A Tool for Visualizing and Post-processing Deep Learning Anomaly Detection Results for Industrial Time-Series\\
}

\author{\IEEEauthorblockN{Błażej Leporowski}
\IEEEauthorblockA{\textit{Dept. of Electrical and Computer Engineering} \\
\textit{Aarhus University}\\
Aarhus, Denmark \\
bl@ece.au.dk}
\and
\IEEEauthorblockN{Casper Hansen}
\IEEEauthorblockA{\textit{Technicon ApS} \\
Hobro, Denmark \\
cha@technicon.dk}
\and
\IEEEauthorblockN{Alexandros Iosifidis}
\IEEEauthorblockA{\textit{Dept. of Electrical and Computer Engineering} \\
\textit{Aarhus University}\\
Aarhus, Denmark \\
ai@ece.au.dk}
}

\maketitle

\begin{abstract}
Industrial processes are monitored by a large number of various sensors that produce time-series data.
Deep Learning offers a possibility to create anomaly detection methods that can aid in preventing malfunctions and increasing efficiency.
But creating such a solution can be a complicated task, with factors such as inference speed, amount of available data, number of sensors, and many more, influencing the feasibility of such implementation.
We introduce the DeTAVIZ interface, which is a web browser based visualization tool for quick exploration and assessment of feasibility of DL based anomaly detection in a given problem.
Provided with a pool of pretrained models and simulation results, DeTAVIZ allows the user to easily and quickly iterate through multiple post processing options and compare different models, and allows for manual optimisation towards a chosen metric.
\end{abstract}

\section{Introduction}
\input{Introduction}

\section{Related work}
\input{Related}

\section{The DeTAVIZ tool}
\input{Method}

\section{Illustrative examples}
\input{Experiments}

\section{Conclusions}
\input{Conclusions}

\section*{Acknowledgment}  \label{S:Acknowledgment}
This work is supported by the Smart Industry project (Grant No. RFM-17-0020) granted by the EU Regional Development Fund.

\bibliographystyle{IEEEtranN}

\bibliography{references}

\end{document}

%% file: Introduction.tex
The problem of anomaly detection in time-series is well represented by various industrial case studies.
In~\cite{Carletti2019}, authors present the application of anomaly detection using DL in a refrigerator manufacturing process.
Authors of~\cite{theumer2021anomaly} propose and test a method to detect anomalies in industrial time-series with the aim of retaining energy efficiency.
In \cite{Canizo2019} and \cite{Liang2021} authors propose methods for multivariate time-series and assess them against real life industrial data.
Authors of \cite{Chen2020}  introduce an unsupervised anomaly detection method based on sliding windows for industrial robots, which can perform real-time anomaly detection on multivariate time-series data. The authors further verify the method on an industrial robot.

Deep Learning (DL) has the potential to aid a variety of manufacturing processes by improving the anomaly detection and prediction.
Establishing whether the results of a DL solution will be useful in a specific case, however, can be a complex task.
Often the DL aided anomaly detection is an addition to an already existing process, and because of that opportunities for experimentation can be limited.

This paper presents the DeepTimeAnomalyViz (DeTAVIZ), an interface for visualising and exploring results of anomaly detection simulations on time-series datasets.
DeTAVIZ takes the model classification or prediction series as an input, and allows the user to explore and visualize the effects that post-processing those predictions have on the system's anomaly detection performance.
The DeTAVIZ works together with a pool of trained models and saved analysis results, and it will automatically pick the performing model and its simulation results based on the user's choice of dataset parameters, such as data dimensionality or other characteristics.

The presented method allows to quickly assess and compare viability of different models and dataset characteristics for the real-time anomaly detection task.
This solution can provide a user-friendly, low manual effort, and rapid first step in the process of enhancing a manufacturing process with DL based anomaly detection.

%% file: Related.tex
In \cite{Paraschos2021}, the authors propose VisioRed, a visualisation tool for interpretable predictive maintenance.
This system is compatible with multi-variate time-series data, and its end product is the prediction of a component's remaining useful lifetime (RUL).
Similarly to DeTAVIZ, VisioRed allows the user to explore what-if scenarios and explain the decisions that the model made.
The tool is thus related to DeTAVIZ in its look and interactivity, but targets a different use case and offers different functionality.

VisPlause \cite{Arbesser2017} is a system designed to support an efficient inspection of data quality problems in time-series.
It is a complex tool that aims to determine data quality problems by applying flexible plausibility checks and utilizing the time-series meta data.
These characteristics make it specialized and difficult to generalize into different datasets.
VisPlause is meant to operate on raw time-series inputs, as part of the data pre-processing stage, while DeTAVIZ operates on the classification or prediction results obtained by using ML models.

%% file: Method.tex
\begin{figure*}[]
    \begin{center}
        \includegraphics[width=\textwidth]{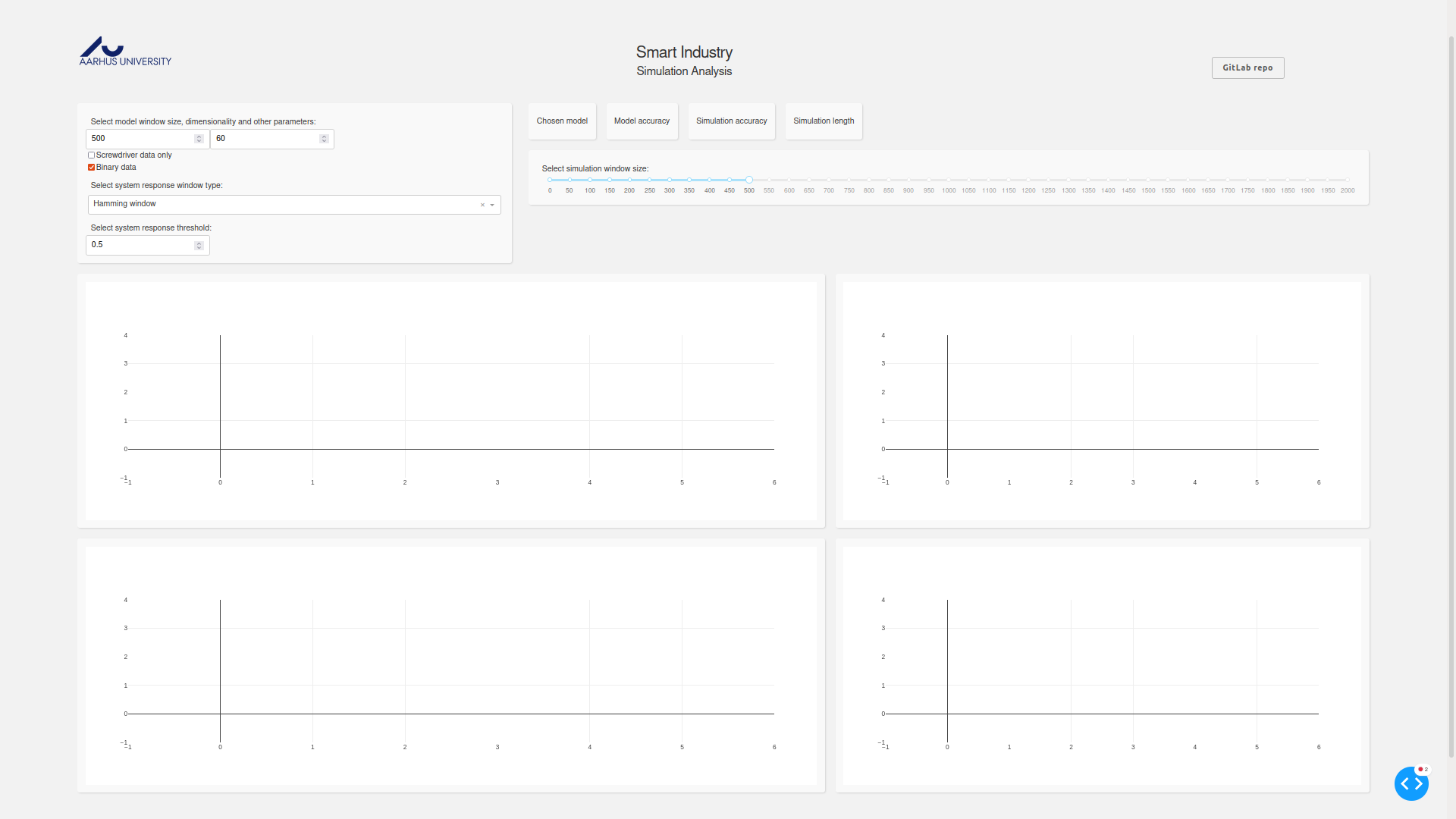}
    \end{center}
    \caption{DeTAVIZ web interface.}
    \label{fig:empty_interface}
\end{figure*}

The main motivation behind DeTAVIZ is to allow the user to determine, in a simple and intuitive way, whether any of the multiple models trained on different variants of the same dataset present a promising solution to the anomaly detection problem in a the manufacturing process at hand.
The visualisation is based on analysis results.
Each analysis is obtained by a unique combination of a specific model and dataset characteristic, such as data dimensionality or sliding window size.
The analysis results consist of a predicted label, ground truth and run time for each sample, operating at a time-instance level of the input time-series data.

Upon opening the DeTAVIZ, the user is prompted to select the model's window size, dimensionality and other parameters.
The tool then proceeds to search the pool of available trained models, selects a model with the highest test average F1 score, and then it loads the analysis results for this model.
The model and analysis basic characteristics, like model name, accuracy and analysis length are displayed.

\begin{figure*}[]
    \begin{center}
        \includegraphics[width=\textwidth]{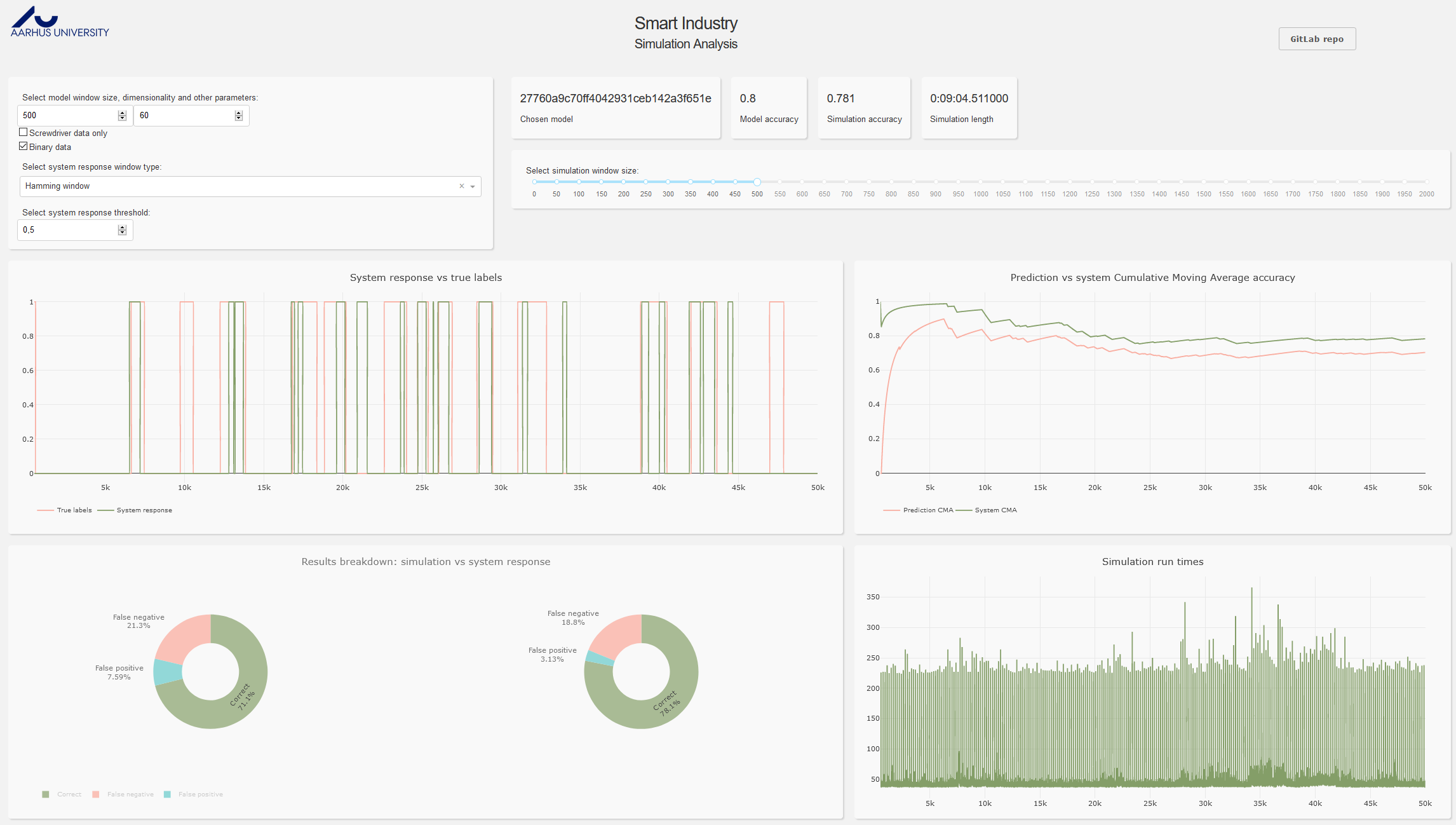}
    \end{center}
    \caption{Example case of a screwdriving anomaly detection coming from the AURSAD dataset.}
    \label{fig:web_interface}
\end{figure*}

\begin{figure}[]
    \begin{center}
        \includegraphics[width=0.55\linewidth]{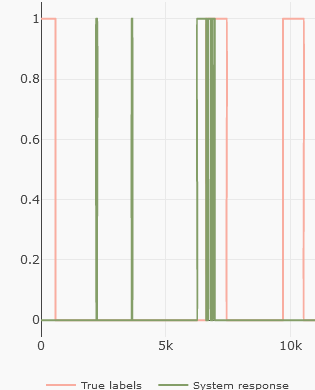}
    \end{center}
    \caption{Noisy model predictions corresponding to the raw DL model predictions.}
    \label{fig:noisy_response}
\end{figure}

\begin{figure}[]
    \begin{center}
        \includegraphics[width=0.55\linewidth]{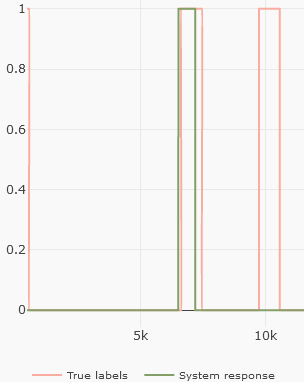}
    \end{center}
    \caption{Post-processed time-series anomaly detection response obtained by using appropriate window type, size and threshold values.}
    \label{fig:smoothed_response}
\end{figure}

The interface of DeTAVIZ is illustrated in Figure \ref{fig:empty_interface}. 
The user needs to specify both the model parameters and the post-processing analysis parameters on the panel placed at the top-left of the interface, together with the window size slider to the right.
From the top, the first four text fields and checkboxes describe the model parameters. 
The first textbox defines the dataset's window size, an the next textbox defines the dataset's diemnsionality.
Beneath them, there are two checkboxes responsible for declaring whether to load a model trained on a limited selection of dataset features (e.g., as in our experiments focused on screwdriving anomaly detection, to screwdriver only), and whether to load a model trained on a binary data, that is a dataset containing only 2 classes: normal and anomalous, without discerning the types of anomalies. 
The web interface presents 4 interactive plots to the user, as seen in Figure \ref{fig:web_interface}.
A plot of the labels predicted by the system compared to the ground truth labels and a Cumulative Moving Average accuracy in the top row.
In binary mode, label 1 means anomaly, label 0 means normal sample.
In the bottom row, one plot presents simulation run times for each simulation step.

The goal of this plot is to present the user a quick way to assess whether the current model is able to be used in real-time scenarios.
The last plot is the results breakdown double pie chart, where the statistics about the correct, false positive and false negative predictions are presented.

When a model is tasked with predicting the labels one by one for a sequence that consists of a few thousand samples, the result is often noisy, as illustrated in Figure \ref{fig:noisy_response}.
This is a typical case appearing in industrial time-series anomaly detection, where the model generally predicts the anomaly correctly, however, large amount of signal noise can lead the system to produce a significant number of false positives and false negatives.
If such anomaly detection process is to be embedded in some anomaly prevention system operating in real-time, in which the next steps are reliant on the model's predictions, such a result can prove detrimental.
By treating the model's predictions as a new system signal, and applying signal processing techniques to it, the system response can be post-processed at a specific case base, which in turn can increase the overall accuracy and decrease the number of false positives and negatives.

DeTAVIZ provides a thresholded rolling window average as a mean to achieve this goal. 
A moving average is a type of a Finite Impulse Response (FIR) filter, and is one of the most popular tools for smoothing local fluctuations~\cite{guinon2007moving}, \cite{alvarez2005detrending}.
The user can choose the window type and size, as well as the threshold.
The post-processed value for each sample is calculated as a mean of the chosen window type, and then if the value exceeds the threshold it is labeled an anomaly.
The user can experiment by using various choices for window type, size and threshold, leading to different post-processing results, and select those combinations resulting to improved system response, as demonstrated in Figure \ref{fig:smoothed_response}.

\begin{figure*}[]
    \begin{center}
        \includegraphics[width=\textwidth]{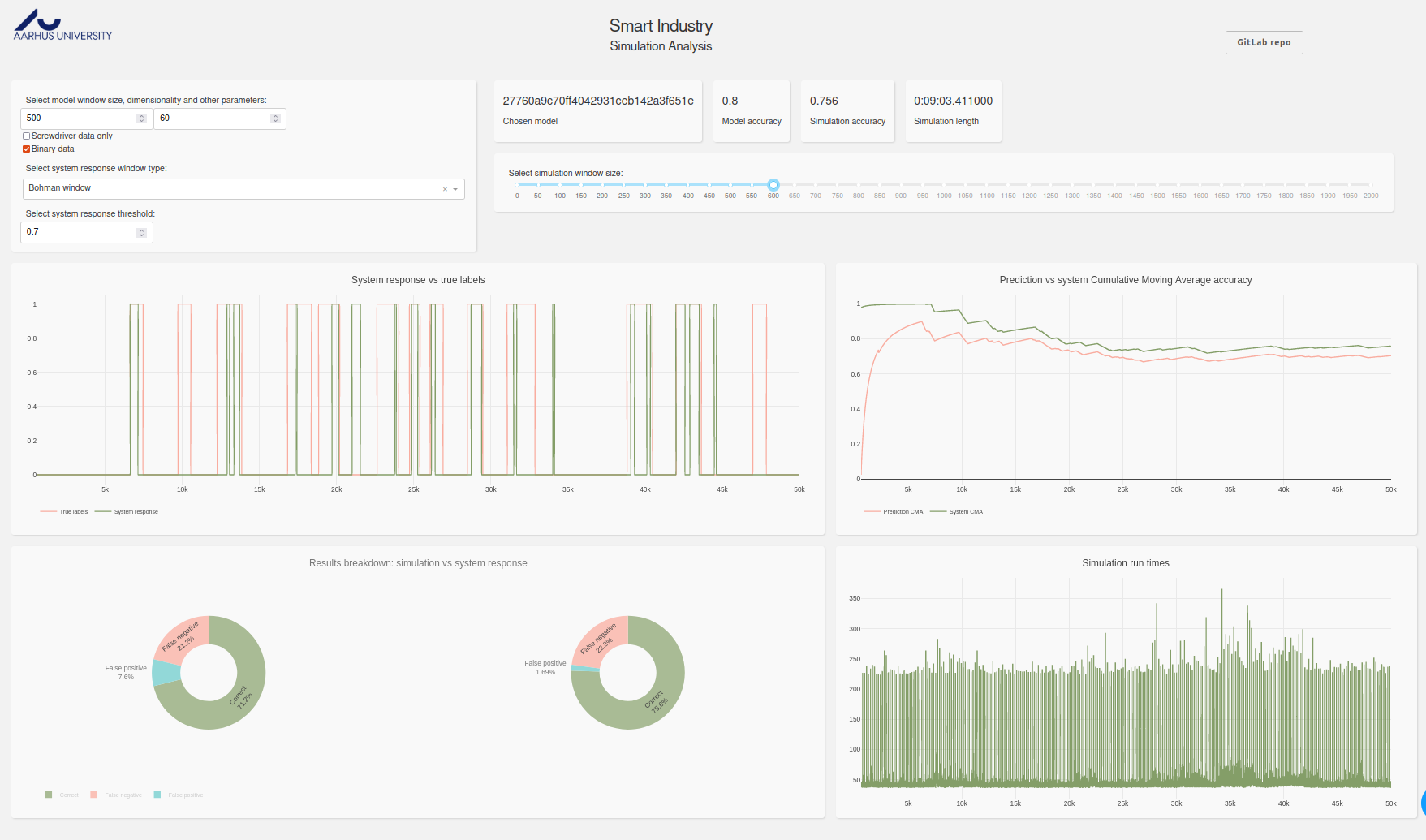}
    \end{center}
    \caption{The same model as in Figure \ref{fig:web_interface} optimised for minimal false positives ratio.}
    \label{fig:minimal_positives}
\end{figure*}

%% file: Experiments.tex

To illustrate the functionalities provided by the DeTAVIZ, we conducted extensive anomaly detection analysis on the recently introduced AURSAD dataset \cite{leporowski2021detecting}.
The AURSAD dataset contains 2,045 samples of robot screwdriving operation each corresponding to normal operation or an anomaly belonging to 4 different categories.
Detailed description of the dataset characteristics and the data collection process can be found in \cite{aursad_arxiv}. 
To facilitate AURSAD dataset pre-processing and easy integration into deep learning experimentation pipelines, the AURSAD python library was also released~\footnote{\url{https://pypi.org/project/aursad/}}.

For the experiments, the binary labeling has been used, i.e., we consider all anomalies to form one class which is to be distinguished from the normal operation, and multiple TABL \cite{Tran2017} and ResNet \cite{resnet} time-series classification models were trained as binary classifiers.
We chose TABL and ResNet models, as they have been shown to provide state-of-the-art results in different types of time-series classification problems.
Our experiments include 43 models trained with different combinations of window size, dimensionality and model hyperparameters.
The user interface, with access to the pre-trained DL models used in our experiments can be found on GitHub\footnote{\url{https://github.com/CptPirx/detaviz}}.

Model 1 is trained on a window size of 500 samples and 60 dimensions.
In the example illustrated in Figure \ref{fig:web_interface}, it can be seen that post-processing the DL model's responses leads to a smooth time-series anomaly detection result, with a low amount of noise.
By applying a rolling mean filter with Hamming window of size 500 with a threshold of 0.5, the system's response contains less sudden changes. 
This, in turn, leads to over two times smaller number of false positive predictions.
The wide assortment of window types together with thresholding allows to optimise towards different goals.
The same model, optimised towards the minimal false positives (FP) ratio, can achieve a false positive ratio of $1.69\%$ using a Bohman window with a size of 600 samples and 0.7 threshold.
Compared to raw predictions, this results in only a slight increase in false negatives and still improving the overall accuracy of the system.
When optimising for false negatives (FN), a rectangular window of size 800 and a threshold of 0.2 leads to decrease of false negatives from 21.3\% in raw data to 8.8 \%, and also improving the overall accuracy.
The results for this configuration are presented in Figure \ref{fig:minimal_positives}.

The same experiments performed on a different model, named further model 2, trained on a window size of 200 samples and 60 dimensions, showed similar improvements. 
Application of a rolling mean filter with a rectangular window of size 200 with a threshold of 0.3 results in a smoothed signal and improved accuracy.
When optimising for false positives, filtering the signal using a Hamming window of size 350 with a threshold of 0.75 provides good improvement while maintaining the overall accuracy.
An improved FN ratio can be achieved achieved by applying a rectangular window of size 400 with a threshold of 0.15. 

\begin{table}[h]
\centering
\caption{Simulation performance for model 1}
\begin{tabular}{@{}llll@{}}
    \toprule
    Response        & Accuracy  & False positives   & False negatives  \\ \midrule
    Raw             & 71.1\%    & 7.6 \%            & 21.3\%           \\ 
    Processed       & 78.1\%    & 3.1 \%            & 18.8\%           \\
    FP optimised    & 75.6\%    & 1.7 \%            & 22.8\%           \\
    FN optimised    & 77.9\%    & 13.3\%            & 8.8 \%           \\
    \bottomrule
\end{tabular}
\label{tab:simulation_results_1}
\end{table}

\begin{table}[h]
\centering
\caption{Simulation performance for model 2}
\begin{tabular}{@{}llll@{}}
    \toprule
    Response        & Accuracy  & False positives   & False negatives  \\ \midrule
    Raw             & 77.3\%    & 4.3 \%            & 18.4\%           \\ 
    Processed       & 79.3\%    & 5.6 \%            & 15.1\%           \\
    FP optimised    & 75.5\%    & 1.2 \%            & 23.3\%           \\
    FN optimised    & 78  \%    & 11.3\%            & 10.7\%           \\
    \bottomrule
\end{tabular}
\label{tab:simulation_results_2}
\end{table}

Tables \ref{tab:simulation_results_1} and \ref{tab:simulation_results_2} provide quantitative comparison between the anomaly detection performance achieved using two DL models responses compared to applying the post-processing illustrated in Figure \ref{fig:web_interface}, as well as manually optimising the models performance towards the least amount of false positives and false negatives. 

As can be seen, by using DeTAVIZ one can improve performance in a user-friendly interface.
The time it takes for the tool to update all plots after changing either filter or model parameters is under 4 seconds, which enables the user to rapidly assess multiple combinations for optimal performance.

%% file: Conclusions.tex
We have presented and showcased the DeTAVIZ tool for visualising and post-processing time-series anomaly detection results provided by DL models on the AURSAD dataset designed for industrial time-series anomaly detection.
Our experiments show that the DeTAVIZ interface, by incorporating signal processing techniques and easy to read visualisations, can lead to significant performance improvements while not requiring coding nor machine learning knowledge. 

Directions of future improvement include: i) make DeTAVIZ dataset agnostic, and ii) incorporate more signal processing functionalities into the tool iii) introduce an automated parameter search to optimise towards a specified metric. 
The principles and visualisations already implemented are well suited for any time-series anomaly detection analysis, and by adding an automated dataset features inference, DeTAVIZ could be used as an intuitive, out of the box tool for quickly assessing the feasibility of DL based anomaly detection in a variety of situations.
Signal filtering and smoothening is an extremely well researched area, with many solutions, ranging from fairly simple, like the one showcased here, to complex methods able to tackle difficult cases.
Thus, by implementing more of the leading solutions for signal filtering, DeTAViz would provide a wide array of proven and effective methods, suitable for different dataasets and DL model's output signal characteristic.